%% file: main.tex
\documentclass[lettersize,journal]{IEEEtran}
\usepackage{amsmath,amsfonts}
\usepackage{algorithmic}
\usepackage{algorithm}
\usepackage{array}
\usepackage{tabu,multirow}
\usepackage[table]{xcolor}
\usepackage{color,colortbl}
\definecolor{Gray}{gray}{0.9}
\usepackage{booktabs,makecell}
\usepackage[caption=false,font=normalsize,labelfont=sf,textfont=sf]{subfig}
\usepackage{textcomp}
\usepackage{stfloats}
\usepackage{url}
\usepackage{verbatim}
\usepackage{graphicx}
\usepackage{cite}
\usepackage{caption}
\definecolor{Gray}{gray}{0.9}
\newcommand{\refFig}[1]{Fig. \ref{#1}}
\newcommand{\refTab}[1]{Tab. \ref{#1}}

\begin{document}

\title{OpenSlot: Mixed Open-Set Recognition with Object-Centric Learning}

\author{Xu Yin, Fei Pan, Guoyuan An, Yuchi Huo, Zixuan Xie, Sung-Eui Yoon\thanks{This work was supported by the Institute of Information \& Communications Technology Planning \& Evaluation (IITP) grant funded by the Korea government (MSIT) (RS-2023-00237965, Recognition, Action and Interaction Algorithms for Open-world Robot Service), the National Research Foundation of Korea (NRF) grant funded by the Korea government (MSIT) (No. RS-2023-00208506 (2024)), and was partially supported by the National Key R\&D Program of China (No. 2024YDLN0011) and NSFC (No. 62441205) (Corresponding author: Sung-eui Yoon).

Xu Yin and Guoyuan An are with the School of Computing, Korea Advanced Institute of Science and Technology, Daejeon 34141, South Korea (E-mail: yinofsgvr@kaist.ac.kr and anguoyuan@kaist.ac.kr).

Fei Pan is with the School of Computer Science Engineering, University of Michigan. Email: feipan@umich.edu).

Yuchi Huo is with the State Key Lab of CAD and CG, Zhejiang University, China and Zhejiang Lab, China 310058 (E-mail: huo.yuchi.sc@gmail.com).

Zixuan Xie is with the institute of Computing Technology, University of Chinese Academy of Sciences, Beijing, China 100190 (E-mail: xiezixuan211@mails.ucas.ac.cn)

Sung-eui Yoon is with the Faculty of School of Computing, Korea Advanced Institute of Science and Technology, Deajeon 34141, South Korea (E-mail:
sungeui@gmail.com).
}}

\markboth{Journal of \LaTeX\ Class Files,~Vol.~14, No.~8, August~2021}%
{Shell \MakeLowercase{\textit{et al.}}: A Sample Article Using IEEEtran.cls for IEEE Journals}

\maketitle

\input{sec/abstract} 
\section{Introduction}
\label{sec:introduction}
\input{sec/introduction}
\section{Related Work}
\label{sec:related_work}
\input{sec/relatedwork}
\section{Mixed Open-set recognition}
\label{sec:verification}
\input{sec/Verification_experiments}
\section{Method}
\label{sec:method}
\input{sec/Method}
\section{Evaluation}
\label{sec:evaluation}
\input{sec/Evaluation}

\section{Conclusion}
\label{sec:Conclusion}
\input{sec/Conclusion}
\appendix
\input{sec/Appendix}

\bibliography{egbib}
\bibliographystyle{IEEEtranS}

\end{document}

%% file: sec/abstract.tex
\begin{abstract}   
Existing open-set recognition (OSR) studies typically assume that each image contains only one class label, with the unknown test set (negative) having a disjoint label space from the known test set (positive), a scenario referred to as full-label shift. This paper introduces the mixed OSR problem, where test images contain multiple class semantics, with both known and unknown classes co-occurring in the negatives, leading to a more complex super-label shift that better reflects real-world scenarios. To tackle this challenge, we propose the OpenSlot framework, based on object-centric learning, which uses slot features to represent diverse class semantics and generate class predictions. The proposed anti-noise slot (ANS) technique helps mitigate the impact of noise (invalid or background) slots during classification training, addressing the semantic misalignment between class predictions and ground truth. We evaluate OpenSlot on both mixed and conventional OSR benchmarks. Without elaborate designs, our method not only excels existing approaches in detecting super-label shifts across OSR tasks, but also achieves state-of-the-art performance on conventional benchmarks. Meanwhile, OpenSlot can localize class objects without using bounding boxes during training, demonstrating competitive performance in open-set object detection and potential for generalization.

\end{abstract}

%% file: sec/introduction.tex
Open-set recognition (OSR) assumes the test samples come from open space~\cite{osr_review,tmm_osr,tmm_open_set}, aiming to reject those having unknown classes (\textit{\textbf{negative}}) and maintaining models' classification performance on known classes (\textit{\textbf{positive}}). This scenario is
 vital to ensure the safety of machine learning systems, and has been widely applied in various fields, e.g., fault detection~\cite{Anomaly_Detection} and medical diagnosis~\cite{medical}, where “unknownnes" may happen anywhere and anytime.

Currently, OSR is categorized as a generalized Out-of-distribution (OOD) detection task~\cite{ood_review}, where the unknown classes are regarded as a label shift relative to the known classes. A standard OSR pipeline is to train a multi-class classification model on the known training images and compute a score~\cite{knn,energy_ood} to measure the test sample's “knownness” and decide whether to reject or accept. 

\begin{figure}[t]
\begin{center}
\includegraphics[width=8.35cm]{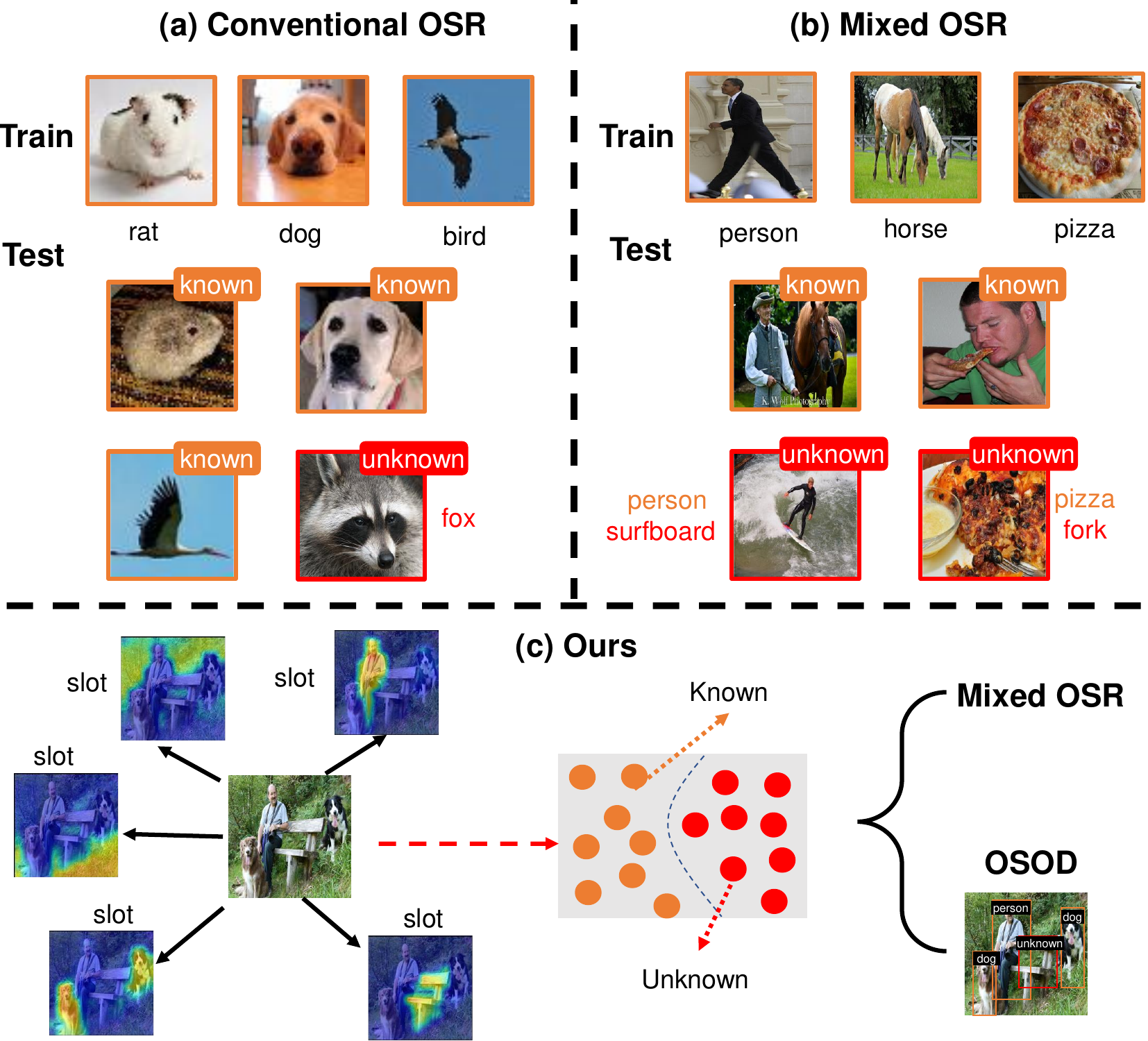}
\end{center}
   \caption{(a) Conventional OSR assumes test samples only contain one class semantic (single-label) and regards images from new classes as unknown.
 (b) We propose the mixed OSR problem, which tests on images with multiple classes. The negative samples contain known and unknown classes and require the classifier to accurately detect the occurrence of unknown classes. (3) Existing methods can easily ignore the unknown class due to the existence of the known content. To address this issue, we propose using slots for semantic representation and producing independent class predictions. Thus, we learn a clear known/unknown decision boundary. Besides, our method can localize the unknown class and realize open-set object detection (OSOD).}
   \label{fig: idea}
\end{figure}

The present OSR framework (\refFig{fig: idea}, (a)) builds upon an ideal condition: every test image is single-label. It leads to biased assumptions: the negative image only contains unknown classes, yielding two completely different label spaces of positive and negative images (termed a full-label shift). 

Nevertheless, real images are often multi-labeled, with mixed known and unknown class objects. In this study, we formulate a \textbf{mixed OSR} problem (\refFig{fig: idea}, (b)) that considers test samples with multiple class semantics, which is more commonly seen in real life. In each negative image, the known and unknown classes occur jointly. In this setting, the class set of negative images is the super-set of positives. Therefore, we define mixed OSR as a super-label shift.

Discriminating images containing different class types requires the classification model to distinguish all object semantics clearly. Existing methods usually aggregate class semantics and compute an overall score for OSR evaluation. This approach has a risk: the classifier can be misled by known classes in the images, ignore the unknown content, and thus make an incorrect decision. Resorting to object detection and semantic segmentation approaches can be a good choice for accurately identifying the unknown class object. However, using the pixel and bounding box labels for training would inevitably increase the computation cost and have generalization concerns~\cite{detection_shift,open_seg}.

Recently, object-centric learning~\cite{ocl} has achieved promising results in multi-object representation~\cite{generalization_ocl,dinosaur}. The slot features are bound with different object semantics and carry various object property information, e.g., size and location,  thus facilitating different downstream vision tasks~\cite{slot_tracking,slot_diffusion}.  

Motivated by these advances~\cite{dinosaur, steve}, we present a novel OpenSlot framework for learning slot features to represent different class semantics and implementing image classification to handle the mixed OSR. To isolate the invalid and suspicious background slots during classification, we propose a simple yet effective anti-noise-slot (ANS) technique to address the semantic misalignment problem of the class prediction. Combined with out-of-distribution (OOD) detection scoring metrics, we measure the “knownness” of the object semantic that each slot corresponds to, and achieve significant improvements on various OSR benchmarks. Additionally, our method could explicitly localize unknown objects without using bounding boxes for training, achieving competitive results in open-set object detection tasks. 

In summary, our contributions are three-fold:
\begin{itemize}
    \item We introduce the mixed OSR, a classification task that requires models to detect unknown class objects in images containing known content.
    \item We develop an OpenSlot framework based on object-centric learning. With the proposed anti-noise-slot (ANS) technique, we exclude the invalid and background semantics captured by slots from classification and address the semantic misalignment of class prediction. 
    \item  Extensive experiments show OpenSlot exceeds prevalent OSR studies considerably on various mixed \& conventional benchmarks. Further, our method achieves meaningful open-set object detection results and demonstrates benefits in computational cost and generalization.  
\end{itemize}

%% file: sec/relatedwork.tex
\noindent\textbf{Open-set recognition (OSR).} OSR targets detecting the semantic novelties (i.e., label shift) in the test samples. Prevalent studies~\cite{arpl,difficulty_simulator} usually assume that negative images have a completely different label space from positives, termed the full-label shift. Popular OSR methods include prototype learning~\cite{cssr} and generative techniques~\cite{difficulty_simulator}. For example, CSSR~\cite{cssr} models each known class prototype with an independent auto-encoder manifold to better measure class belongingness. DIAS~\cite{difficulty_simulator} uses GAN to simulate unknown samples, thus improving the classification training. Despite advancing results in detecting the full-label shift in the conventional single-label classification, the complex super-label shift that occurs in images with multiple class types has never been explored. In this work, we present the mixed OSR problem and propose a novel object-centric learning framework to overcome the displayed semantic misalignment problem.

\noindent\textbf{Object-centric learning (OCL).} OCL~\cite{dinosaur,inductive_ocl,steve} represents image semantics with a fixed number of feature vectors named slots. Each slot binds with an object or denotes an independent meaningful semantic and captures its object properties. Compared with CNN/ViT-based models, empirical studies~\cite{inductive_ocl,generalization_ocl} demonstrated that slot-based representation is generalizable and robust against different distribution shifts. With these intriguing properties, OCL has been applied to various vision tasks, such as image generation~\cite{steve,slot_diffusion}, tracking~\cite{slot_tracking}. Recently,  Dinosaur~\cite{dinosaur} introduced self-supervised feature reconstruction as the objective function and first achieved real-world object-centric representations without using priors.

Motivated by these advances, we introduce OCL to handle the mixed OSR. With the proposed ANS technique, we represent diverse image semantics with slots and adapt slot-based predictions for image classification. Combined with existing out-of-distribution (OOD) detection metrics~\cite{energy_ood,multi_label_ood}, we conduct the slot-based OSR.  

\noindent\textbf{Open-set object detection (OSOD).} OSOD targets localizing unknown objects during inference and shares a similar problem definition as out-of-distribution object detection~\cite{osod_2,vos}. There are two main directions in this field: one line aims to break the limit on the category number and recognize more class objects beyond the closed set. For example, Grounding DINO~\cite{grounding_dino} combines with the power of language models~\cite{osod_2} and proposes a cross-modality fusion solution to fuse text and image features at different phases, thus detecting arbitrary class objects. The other line concentrates on unknown detection for safety purposes. For instance, VOS~\cite{vos} generates virtual outliers in training and uses them to estimate a decision boundary between the known and unknown class objects. However, most existing methods~\cite{siren,grounding_dino} are built upon well-trained object detectors, requiring expensive bounding box labels for training. In this study, we target unknown object detection and extend our classification approach to OSOD, achieving competitive results while offering various benefits.

%% file: sec/Verification_experiments.tex
In this section, we introduce the mixed OSR and illustrate its difference from the conventional OSR problem setting.
 
OSR~\cite{osr_review,ood_review} needs to train a multi-class classifier to 1) classify test samples that fall into “known known classes (KKC)” and 2) detect test samples containing “unknown unknown classes (UUC)”. KKCs indicate the known training categories (denoted with $\mathcal{K}$), and UUCs represent the novel category group (denoted with $\mathcal{A}$) that only occurs during inference. The occurrence of new categories at the test time is considered a label shift regarding the set of $\mathcal{K}$.       

Let $\mathcal{D}_{train}$ and $\mathcal{D}_{test}^{k}$ ($k$: known) be the known training and test set (labeled \textbf{\textit{positive}}), and their label space is $\mathcal{K}$ consistently. We denote the unknown test set with $\mathcal{D}_{test}^{u}$ ($u$: unknown) that contains UUCs (labeled \textbf{\textit{negative}}), and express its label space with $\mathcal{Y}$. At the test time, we require the classifiers to accurately recognize negatives when processing samples from $\mathcal{D}_{test}^{k}$ and $\mathcal{D}_{test}^{u}$.

The conventional OSR problem assumes that each image has a single label and considers no KKCs exist in $\mathcal{D}_{test}^{u}$, i.e., $\mathcal{Y}=\mathcal{A}$. This implies that each negative image $\mathbf{x_{i}}\in \mathcal{D}_{test}^{u}$ only has UUC and $\mathcal{D}_{test}^{u}$ and $\mathcal{D}_{test}^{k}$ have entirely disjoint label spaces, i.e., $\mathcal{Y}\cap\mathcal{K}=\emptyset$; we refer to this setting as \textbf{full-label shift}. While this formulation simplifies the OSR problem by making certain assumptions about the content composition of $\mathbf{x_{i}}$, it may not fully capture the complexities of real-world scenarios where label shifts can occur in more diverse forms.

\noindent\textbf{Mixed OSR:} In an unpredictable open space, we consider the mixed scenario, where the negative image $\mathbf{x_{i}}$ contains different category types, a common occurrence in real-world settings.

In this setting, we use $y_i$ to denote $\mathbf{x_{i}}$'s class label set, and have $y_i\cap \mathcal{K} \neq \emptyset \quad \text{and}\quad y_i\cap \mathcal{A} \neq \emptyset$. Hence, the label space becomes $\mathcal{Y}=\mathcal{A}\cup\mathcal{K}$, and the distribution shift occurs in the superset of $\mathcal{K}$, termed \textbf{super-label shift}.
 
Handling this mixed problem requires OSR models to recognize all class objects within the negative sample $\mathbf{x_{i}}$ and classify them as known or unknown. Furthermore, detecting the super-label shift in images with multiple class semantics is crucial for extending OSR to multi-label scenarios; a larger label space implies more mixed scenarios, with distribution shifts potentially manifesting anywhere. This capability is directly beneficial for real-world applications, such as autonomous driving, species discovery, medical imaging analysis~\cite{medical}, and defect detection~\cite{Anomaly_Detection} in manufacturing processes.

Intuitively, classifying an image with multiple semantics can be influenced by the presence of known class objects. One potential solution is to leverage pre-trained dense predictors, such as object detectors~\cite{faster_rcnn} and segmentation methods~\cite{opengan}, to implement open-set object detection~\cite{osod_2,OSOD_3} and open-set image segmentation~\cite{open_seg}, classifying each object proposal and pixel for more accurate known/unknown identification. However, these dense approaches require extensive labeled data, are computationally expensive (as discussed in Sec. \ref{sec:osod}), and are sensitive to domain shifts~\cite{detection_shift}.

In contrast, our mixed OSR approach relies solely on class labels during training. It perceives label shifts in multi-label and mixed scenarios by learning generalizable representations of diverse semantics within images and establishing clear decision boundaries between known and unknown categories.

%% file: sec/Method.tex
\begin{figure*}[t]
\begin{center}
\includegraphics[width=16cm]
{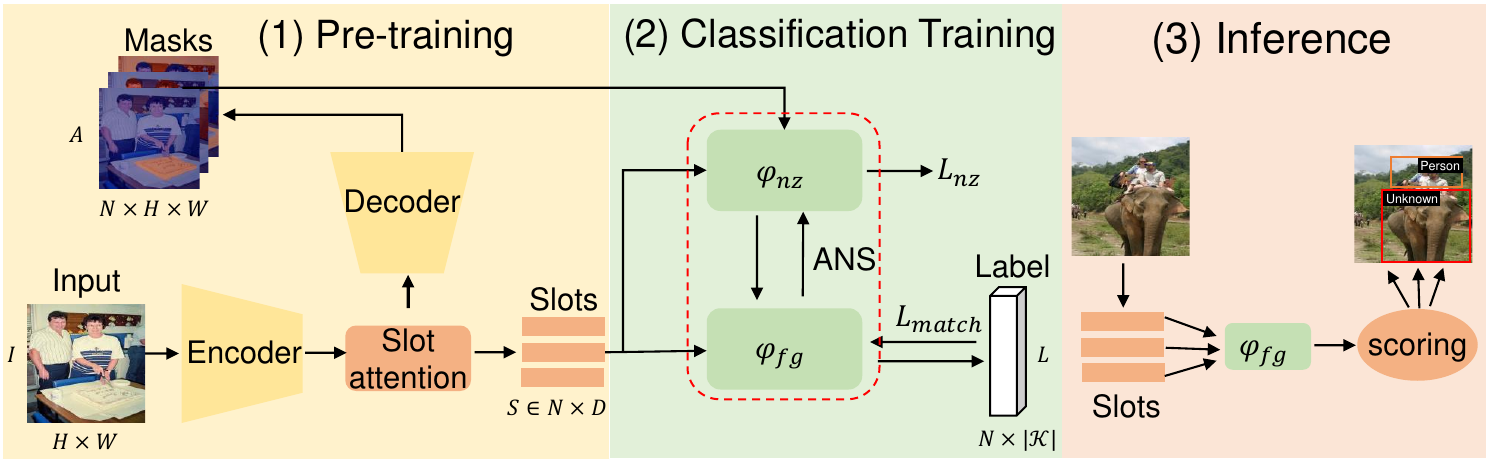}
\end{center}
   \caption{OpenSlot overview. In (1), we pre-train Dinosaur~\cite{dinosaur} (following the encoder-decoder design, the encoder can be either ResNet or Transformer architecture) on the known training images, using slots $S$ to learn class semantics. Next (2), we add two classifiers, the noise classifier $\varphi_{nz}$ and the main classifier $\varphi_{fg}$, trained with $L_{nz}$ (Eq. \ref{eq:nz_loss}) and $L_{match}$ (Eq. \ref{eq:match_loss}) respectively, to identify the noise (invalid and background) slots and implement semantic classification with the proposed ANS technique (\refFig{fig:ANS_presentation}). During inference (3), we score slot-based class predictions to address the mixed OSR.}
   \label{fig:overview}
\end{figure*}

In this part, we first introduce object-centric learning (OCL) and clarify the motivations. Next, we propose an OpenSlot framework to handle the mixed open-set recognition (OSR).  
\subsection{Preliminary: Object-centric learning}
The goal of OCL~\cite{ocl,cvpr_slots} is to represent the visual input with a set of object features (termed slots), with which to obtain the compositional representation of the input to facilitate downstream vision tasks~\cite{slot_tracking,slot_diffusion}. In training, slots are randomly initialized and refined by an iterative attention module~\cite{ocl}, to capture an independent object entity through semantic grouping~\cite{savi++}.

OCL commonly adopts the autoencoder design and uses input reconstruction as the objective function. Recently, Dinosaur~\cite{dinosaur} proposes reconstructing the self-supervised features, e.g., DINO~\cite{Dino}, during training and scales OCL to natural images, achieving the first success in the unsupervised setting. This study also shows that the Transformer decoder~\cite{steve, slate} groups objects of the same class into a single slot, aligning with class-level semantics. Our work chooses Dinosaur because it requires no extra supervision~\cite{savi++} and effectively learns slot-based OCL representations for class semantics, making it well-suited for OCL pretraining and image semantic classification, while leaving other alternative models~\cite{dinosaur_2} for future exploration.

\noindent\textbf{Motivations.} We propose using slot features to capture diverse image semantics and implement class classification for known/unknown discrimination, thus addressing the mixed OSR. Our approach is motivated by two key advantages:

Firstly, each slot represents a distinct object semantic and produces an independent class prediction. When evaluating the “knowness" of a negative image, the slot representing unknown classes (UUCs) contributes a significant score value difference compared to known classes. In contrast, existing methods~\cite{multi_label_ood,closed} typically aggregate all semantics to obtain an overall class prediction. Consequently, the resulting OSR score can be largely influenced by the known classes (KKCs) present in the input (Section~\ref{sec:scoring_comparison} compares our slot-based method with existing studies regarding OSR scoring).
 
 Secondly, slots inherently encode spatial location information~\cite{ocl,dinosaur} of the represented objects, enabling our method to localize unknown object semantics and display the super-label shift of the test sample explicitly. Notably, relying solely on class labels during training, our method can perform open-set object detection (OSOD) and achieves competitive performances (see Sec. \ref{sec:osod}).
\subsection{OpenSlot}
The full framework has three stages. Firstly, we conduct unsupervised pretraining to endow slots with semantic representation ability. Next, we implement the slot-based image classification. With the proposed anti-noise-slot (ANS) technique, we exclude noisy slot features from the classification process, allowing the classifier to focus on true class objects. During inference, we score slot predictions using Out-of-distribution (OOD) detection metrics to make OSR decisions.

\noindent{\textbf{Unsupervised pre-training:}} To begin with, we train Dinosaur~\cite{dinosaur} on the KKC training set $\mathcal{D}_{train}$. 

For each training image $I\in \mathcal{D}_{train}$ ($I\in \mathbb{R}^{ H \times W \times 3}$), we get a set of slots $S=\{S_{1},...S_{N}\}$ and their attention masks $A$ (\refFig{fig:overview} (1)):
\begin{equation}
    \{S, A\}=Dinosaur(I),
    \label{eq:discovery}
\end{equation}
note that $\mathcal{S}\in\mathbb{R}^{ N \times D} $ and $A\in\mathbb{R}^{ N \times H \times W} $, $D$ denotes the slot dimension. Following the setup in ~\cite{dinosaur}, all slots $S_{i}\in S$ are first initialized and randomly sampled from a common distribution. These slots then compete to reconstruct the self-supervised DINO~\cite{Dino} features.

In this work, we mainly adopt the ViT~\cite{vit} encoder and Transformer decoder~\cite{slate,steve}, and $S$ tends to capture category-level semantics. Other designs are compared and discussed in Sec. \ref{sec:evaluation}. Note that regardless of the encoder/decoder architecture, pixels in $A$ denote the importance of the corresponding slot in reconstructing the patch features. 

\noindent{\textbf{Slot-based image classification:}} In addition to semantic representation, slots $S$ capture useful object property information that can benefit downstream prediction tasks~\cite{ocl,generalization_ocl}.
In this section, we leverage the object properties encoded in the slots to perform image classification.

Following the practices in ~\cite{inductive_ocl,generalization_ocl,ocl}, we first pad (with null) the ground-truth categorial vector to make it of the same size as the slot set $S$. Next, we get its one-hot representation $\mathbf{L}\in\mathbb{R}^{N\times|\mathcal{K}|}$ ($|\mathcal{K}|$ denotes the number of training classes).  

A naive solution (we termed Pure Slot) for obtaining class predictions is employing a shared classifier across all slots to generate per-slot predictions. The sum of cross-entropy losses from these predictions is then used as the assignment cost for Hungarian matching~\cite{ocl}. This matching result implements a set-to-set mapping between slot predictions and class labels and is then used to compute the training loss.

However, the number of slots $N$ is fixed during training; OCL learns invalid slots~\cite{ocl} when $N$ is larger than that of practical semantics. Meanwhile, the slots that represent background information also increase the difficulty of matching. These two issues would cause a severe misalignment problem, that is, the final classification decision does not actually represent the class object content of the image. To address these limitations (shown and ablated in Sec. \ref{sec:ans_abl}), we propose the anti-noise-slot (ANS) technique to mitigate the effects of noise (invalid and background) slots on image classification.

\begin{figure}

    \centering
    \includegraphics[width=8.3cm]{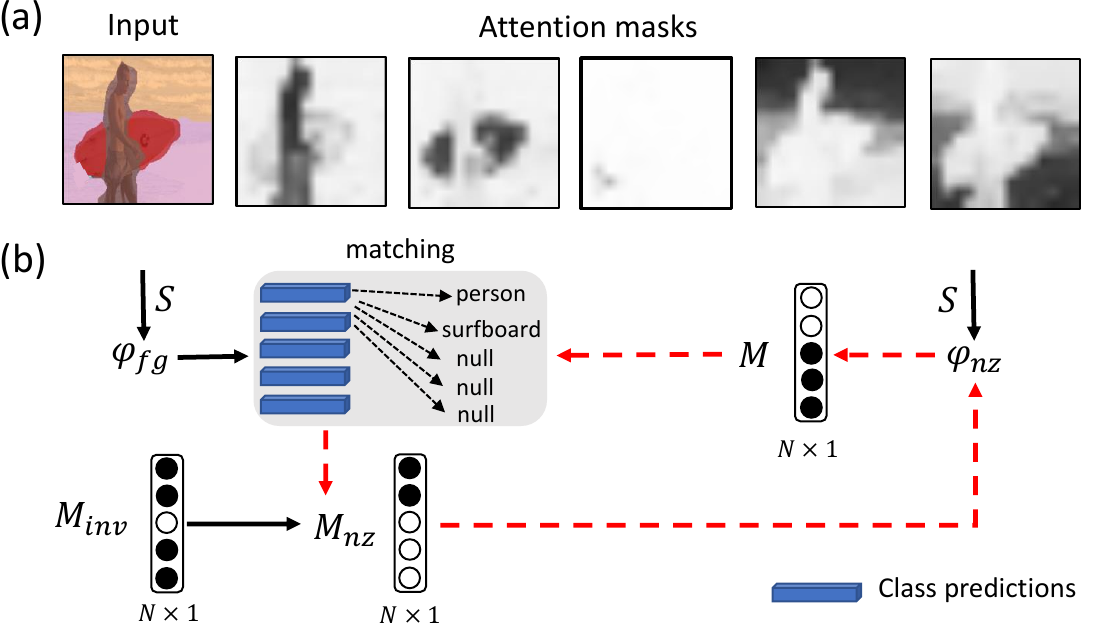}
    \caption{Anti-noise-slot (ANS). After pre-training, we learned slot $S$ to (a) represent the input's diverse semantics (shown in different colors) and get their attention masks (each slot has an attention mask). (b) In each step of the classification training, we: 1) first merge the null slot mask with the invalid slot mask $M_{inv}$ (Eq. \ref{eq:invalid}) to get the pseudo label $M_{nz}$; (2) use $M_{nz}$  to train the noise classifier $\varphi_{nz}$; 3) get the mask $M$ (Eq. \ref{eq:nosiy_mask}) of high-confidence noise prediction from $\varphi_{nz}$ to calibrate (Eq. \ref{eq:match_loss}) the assignment cost of Hungarian matching, enabling  $\varphi_{fg}$ to focus on true class semantics.}
    \label{fig:ANS_presentation}
\end{figure}

Firstly, we add an MLP classifier $\varphi_{fg}$ (shared across slots) after the slot-attention encoder to implement class prediction and matching. To isolate invalid slots, we consider the attention masks $A$ (shown in \refFig{fig:ANS_presentation}) a vital cue that provides rich information for separation. We use a threshold $\alpha$ to determine the semantic region of $A$, and assume that to denote a valid object semantic (foreground/background agnostic), the slot's attention mask should have over two consecutive pixels (corresponding to patch features) over $\alpha$. In this step, we use a mask $M_{inv}\in \mathbb{R}^{N\times1}$ for the separation of the invalid slots:
\begin{equation}
    M_{inv}=\mathbf{1}[A>\alpha],
    \label{eq:invalid}
\end{equation}
Next, we add another shared MLP classifier $\varphi_{nz}$ (shown in \refFig{fig:overview} (2)) to accurately discriminate noise slots. In each classification training step, we get the mask of null slots (not matched with any object class in the label $\mathbf{L}$), and take a union with $M_{inv}$. This way, we build up the pseudo label $M_{nz}\in\mathbb{R}^N$ and train $\varphi_{nz}\in$ with the noise loss $L_{nz}$:
\begin{equation}
    L_{nz}=BCE(\varphi_{nz}(S), M_{nz}),
    \label{eq:nz_loss}
\end{equation}
where BCE is a Binary Cross Entropy loss. Note that the unmatched slots contain suspicious background semantics; we combine them with the invalid slots, and train the noise classifier $\varphi_{nz}$ for accurate discrimination.

To mitigate noise slots' effect on the foreground semantic matching, we get the high-confidence (over a threshold $\beta$, after Min-max normalization) predictions $M\in\mathbb{R}^{N\times1}$ from $\varphi_{nz}$:
\begin{equation}
    M=\mathbf{1}[(\varphi_{nz}(S)>\beta)].
    \label{eq:nosiy_mask}
\end{equation}

Let $L_{set}=CE(\varphi_{fg}(S), \mathbf{L})$ be the pairwise loss between the slot predictions $\varphi_{fg}(S)$ and the label set $\mathbf{L}$, where CE is cross-entropy loss. With the noise mask information from $M$, We first recalibrate the assignment cost, and then apply the Hungarian algorithm to find the matching with the least cost:
\begin{equation}
    L_{match}=HUN((1-M)\cdot L_{set}+\lambda M \cdot L_{set}),
    \label{eq:match_loss}
\end{equation}
$\lambda$ is a large scalar to make noise slots of extremely high assignment costs in the matching process, thus reducing their probability of being matched with the true classes in $\mathbf{L}$.

As shown in \refFig{fig:ANS_presentation}, masks $M$ and $M_{nz}$ are constructed based on the prediction of $\varphi_{nz}$ and $\varphi_{fg}$, respectively. These masks are then used as supervision ($M_{nz}\rightarrow \varphi_{nz}$) and recalibration ($M\rightarrow L_{match}$) of each other's classifiers. This self-supervised technique encourages the main classifier $\varphi_{fg}$ to output low-value logits for noise slots and increases the magnitude difference between noise slots and foreground slots, which represent true class semantics. Meanwhile, the invalid slot mask $M_{inv}$ is constructed heuristicly (with the threshold $\alpha$) and stabilizes the early training of slot classification.

\noindent{\textbf{Framework Overview.}} As illustrated in \refFig{fig:overview}, we firstly pre-train Dinosaur~\cite{dinosaur} on $\mathcal{D}_{train}$ (the training set of known classes) to obtain the representative slot features. Next, we freeze the Dinosaur part and add two classifiers, $\varphi_{nz}$ and $\varphi_{fg}$, following the encoder. After excluding the noise slots from the semantic matching, we implement the image classification. The overall loss is expressed as follows:
\begin{equation}
    L=L_{match}+L_{nz}.
    \label{eq:overall_loss}
\end{equation}
Note that the encoder and decoder part is frozen during the classification training, and all computation overhead comes from the classifiers $\varphi_{nz}$ and $\varphi_{fg}$.

\noindent\textbf{Inference.} For the closed-set classification, we use the max-logit slot in single-label experiments and the maximum class-wise logit across slots in multi-label experiments to determine the semantic classification result.

For OSR tasks, we score slot predictions using standard out-of-distribution (OOD) detection functions~\cite{energy_ood,m_score,msp,odin-score}. These scores measure the degree of label shift for each slot's associated semantic content. The final OSR decision score is obtained by summing the individual scores.
 
OpenSlot can also localize objects to realize open-set object detection, thus explicitly displaying the label shift. We treat attention masks as object proposals and use the over-threshold regions to produce the bounding box results. Based on corresponding OSR scores, we distinguish close-set and open-set objects and assign the known classes and “unknown” labels.

%% file: sec/Evaluation.tex
\label{sec:evaluation}
\subsection{Experiment setup}
In this section, we first outline the experimental setup for different OSR tasks and show OpenSlot's effectiveness. We also conduct extensive ablation studies to validate the impact of our design and compare OpenSlot with existing studies. 

\noindent\textbf{Benchmarks \& Experiments.} 
\begin{itemize}
    \item {\textbf{Mixed OSR:}} We mainly test on Pascal VOC (VOC)~\cite{pascal} and MS COCO (COCO)~\cite{coco} datasets. To depict OSR tasks, we use the values before and after the slash to denote the number of known known classes (KKCs) and unknown unknown classes (UUCs), respectively. Our benchmarks have two types: single-label and multi-label, indicating whether the classification model is trained on \textbf{single-label} or \textbf{multi-label} known training sets.

For single-label benchmarks, we select images with only one class label from two datasets for training, constructing VOC-6/14 and COCO-20/60, respectively. In the multi-label cases, where each training image has more than one labeled class, we build four benchmarks: VOC-COCO (20/60) and COCO-{20/60, 40/40, 60/20}. VOC-COCO takes VOC classes as KKCs and the non-overlapping 60 COCO classes as UUCs. For the other three benchmarks, we randomly select 20, 40, and 60 COCO classes as KKCs, respectively, and use the remaining COCO classes as UUCs.

After the classification training, we evaluate models on test images with multiple class semantics. Throughout all experiments, we construct two types of unknown test sets $\mathcal{D}_{test}^{u}$: the homogeneous ($\mathcal{H}$, where each image contains only UUCs) and the mixed ($\mathcal{M}$, where UUCs are mixed with KKCs). This comprehensive setup evaluates the OSR models' ability to detect the full-label and super-label shifts. Note that in the multi-label setting, we have larger test sets and more complex mixture cases, which evaluate OSR models' robustness.  Detailed class splits and benchmark information are provided in the Appendix.
    
\item {\textbf{Conventional OSR:}} We also evaluate OpenSlot on three conventional OSR benchmarks: CIFAR-6/4~\cite{CIFAR10}, CIFAR-50/50, and TinyImageNet-(20/180)\cite{tinyimagenet}. To ensure a fair comparison with existing studies, we adopt the official KKC/UUC splits from the OpenOOD\cite{openood} benchmark.
\end{itemize}

\noindent\textbf{Baselines \& Metrics.} For single-label tasks of the mixed OSR, we compare OpenSlot with state-of-the-art (SOTA) methods. Since there are no specific OSR studies for the multi-label setting, we use ResNet50~\cite{resnet} for classification training and score predictions with out-of-distribution (OOD) detection functions~\cite{multi_label_ood,m_score,msp} to build baselines for comparison.

For the conventional OSR tasks, we follow protocols in~\cite{openood, vos} and compare our OpenSlot with existing studies~\cite{knn,difficulty_simulator}. We evaluate all results with two metrics: 1) the threshold-free area under the Receiver-Operator curve (AUROC), and 2) the false-positive rate at 95\% true-positive rate (FPR@95).

\noindent\textbf{Implementation details.} In the pre-training on VOC and COCO, we follow the configurations in~\cite{dinosaur}, setting the number of slots to 6 and 7, respectively. For experiments in conventional OSR benchmarks~\cite{openood}, we set the slot number to 5. After the classification training,  we obtain 79.0/74.2 Acc in the closed sets of single-label VOC/COCO and 71.2/65.6 mAP in the multi-label cases, respectively.

\subsection{Results of OSR}

This part reports the mixed OSR results on the constructed single \& multi-label benchmarks. Additionally, we evaluate the performance of OpenSlot on conventional OSR datasets.

\noindent{\textbf{Single-label.}} \refTab{tab:single_label_osr} compares OpenSlot with existing OSR methods, including CSSR~\cite{cssr}, ARPL~\cite{arpl}, MLS~\cite{closed}, and a transformer baseline VIT-MLS~\cite{closed} First, all baselines experience a substantial accuracy degradation on the mixed test set $\mathcal{M}$ (compared to $\mathcal{H}$). This indicates that the super-label shift in the mixed OSR is more challenging than the conventional full-label shift, revealing the existing models' limitations in processing mixed semantics.

In contrast, OpenSlot outperforms all baselines across benchmarks. Specifically, on the $\mathcal{M}$ sets of the two benchmarks, we obtain improvements of 4.1\% and 1.1\% in AUROC and reductions of 1.4\% and 1.8\% in FPR@95 compared to the best baseline results. This demonstrates that OpenSlot can better handle images with mixed class types and distinguish between known and unknown class objects. 

\begin{table}
\scriptsize
    \centering
    \begin{tabular}[t]{c<{\centering}|c<{\centering} c<{\centering} | c<{\centering} c<{\centering}}
    \hline
    \multirow{3}{*}{\textbf{Method}}&\multicolumn{2}{c|}{\textbf{VOC-6/14}}&\multicolumn{2}{c}{\textbf{COCO-20/60}}\\
    ~&$\mathcal{H}$& $\mathcal{M}$&$\mathcal{H}$& $\mathcal{M}$\\
    ~&\multicolumn{2}{c|}{AUROC $\uparrow$  / FPR@95 $\downarrow$}&\multicolumn{2}{c}{AUROC $\uparrow$ / FPR@95 $\downarrow$}\\
    \hline
    CSSR~\cite{cssr} &70.1 / 95.8& 53.8 / 96.2&68.7  / 91.9&56.1 / 97.0\\
    MLS~\cite{closed}&63.5 / 91.3& 55.9 / 93.3 &76.3  / 86.8&67.8 / 90.1\\
    ARPL~\cite{arpl} &62.5 / 93.1&56.0 / 92.0&77.8 / 84.5&68.1 / 89.0\\
        DIAS~\cite{difficulty_simulator} &62.4 / 93.5& 53.6 / 93.2&65.5 / 92.1&61.4 / 92.9\\
                ViT-MLS &92.1 / 56.6& 67.2 / 88.1&90.6 / 69.2&73.8 / 84.9\\
    \hline
   \rowcolor{Gray}OpenSlot&\textbf{93.2 / 53.3}&\textbf{71.3 / 86.7}&\textbf{91.6 / 70.1}&\textbf{74.9 / 83.1}\\
    \hline
            \end{tabular}
            \caption{\textbf{Single-label}. In this paper, we utilize \textbf{$\mathcal{H}$} and $\mathcal{M}$ to refer to the UUC-only test set and the mixed test sets that contain both KKCs and UUCs. Our results, scored using the energy function~\cite{energy_ood}, consistently outperform existing studies.
            }
            \label{tab:single_label_osr}
\end{table}

\begin{table*}
\small
    \centering
    \begin{tabular}[t]{
    c<{\centering}|c<{\centering} c<{\centering} |c<{\centering}   c<{\centering}  c<{\centering} c<{\centering} c<{\centering} c<{\centering} }
    \hline
    \multirow{4}{*}{\textbf{Method}}&\multicolumn{2}{c|}{\textbf{VOC-COCO (20/60)}}&\multicolumn{6}{c}{\textbf{COCO}}\\
    ~&\multirow{2}{*}{$\mathcal{H}$}&\multirow{2}{*}{$\mathcal{M}$}& \multicolumn{2}{c}{Task1 (20/60)}& \multicolumn{2}{c}{Task2 (40/40)}& \multicolumn{2}{c}{Task3 (60/20)}\\
    ~&~&~&$\mathcal{H}$&$\mathcal{M}$&$\mathcal{H}$&$\mathcal{M}$&$\mathcal{H}$&$\mathcal{M}$\\
        ~&\multicolumn{2}{c|}{AUROC $\uparrow$ / FPR@95 $\downarrow$}&\multicolumn{2}{c}{AUROC $\uparrow$ / FPR@95 $\downarrow$}&\multicolumn{2}{c}{AUROC $\uparrow$ / FPR@95 $\downarrow$}&\multicolumn{2}{c}{AUROC $\uparrow$ / FPR@95 $\downarrow$}\\
    \hline
    Mahalanobis ~\cite{m_score} &80.3 / 56.7& 61.2 / 87.8&82.3 / 70.3&74.0 / 70.3&72.5 / 79.2&63.3 / 92.3&74.0 / 78.1&64.1 / 91.5\\
    MaxLogit~\cite{closed}&53.5 / 96.0&37.0 / 99.6&39.3 / 98.9&28.4 / 99.2&41.4 / 94.8&39.2 / 96.9&49.2 / 96.6 & 32.2 / 98.7\\
    Odin~\cite{odin-score} &67.3 / 59.8& 42.7 /  92.7&69.4 / 59.1&44.4 / 93.3&74.8 / 53.3&38.6 / 93.2&83.3 / 50.1&40.7 / 93.3\\
        JointEnergy~\cite{multi_label_ood} &85.9 / 58.6& 47.9 / 91.8&92.0 / 48.1&57.7 / 89.7&91.0 / 53.4&58.1 / 92.0&92.7 / 46.0&58.4 / 90.9\\
    \hline
               \rowcolor{Gray} OpenSlot (ResNet34)&\underline{84.7} / 64.8 & 71.5 / 85.3&82.7 / 75.4&75.9 / 81.5&\underline{84.1} / 69.2&64.2 / 91.4 & 83.9 / 72.6 & 69.3 / 90.8 \\
           \rowcolor{Gray} OpenSlot (ViT-B16)&84.4 / 65.0 & \textbf{70.0 / 84.3}&\underline{84.8} / 74.7&\textbf{78.0} / \underline{82.3}&83.7 / 69.1&\textbf{67.5} / \textbf{90.6} & \underline{84.5} / 69.5 & \textbf{70.3} / \textbf{89.1} \\
            \hline
            \end{tabular}
            \caption{\textbf{Multi-label} OSR. Values in brackets are the number of KKCs/UUCs. Bold and underlined indicate the best and the second-best results by ours (with the energy~\cite{energy_ood}). We test OpenSlot using ResNet34~\cite{resnet} and ViT-B16~\cite{vit} as the encoder for training and report their OSR results, respectively.
            }
            \label{tab:multi_label_osr}
\end{table*}
\noindent{\textbf{Multi-label.}} 
\refTab{tab:multi_label_osr} reports our comparison results with different OOD detection methods (working with ResNet50). For a fair comparison, we also train OpenSlot with the ResNet34 encoder and MLP decoder design~\cite{dinosaur} and conduct OSR evaluation. On the $\mathcal{H}$ (full-label shift) test set, OpenSlot falls behind the leading baseline, JointEnergy~\cite{multi_label_ood}, though it remains competitive with others.

The comparison on the $\mathcal{M}$ set reveals that the ViT encoder exhibits a slight performance advantage over ResNet34, and OpenSlot significantly exceeds the baselines regardless of the encoder choice. These observations suggest that our model is robust in handling mixed scenarios, demonstrating strong potential in real-world applications, e.g., detecting unknown obstacles in robotic environments or identifying unseen lesions or abnormalities within the same medical image.

\noindent{\textbf{Conventional OSR.}}
We report the comparison results in \refTab{tab:conventional_osr_results}. On the homogeneous unknown test set $\mathcal{H}$ (indicating the full-label shift), our method outperforms all baselines across three benchmarks, achieving state-of-the-art performances on these conventional datasets. This remarkable performance demonstrates the advantages of our proposed approach in detecting the full-label shift against conventional OSR methods.

\begin{table}[t]
\small
    \centering
    \begin{tabular}{c |c c c  }
    \hline
        Method &CIFAR-6/4&CIFAR--50/50 & Tiny-20/180\\
         \hline
         ODIN~\cite{odin-score} &72.1&80.3&75.7\\
         MSP~\cite{msp} &85.3&81.0&73.0\\
         MLS~\cite{closed}  &84.8&82.7&75.5\\
         KNN~\cite{knn} &86.9&83.4&74.1\\
         \rowcolor{Gray}OpenSlot &\textbf{87.4}&\textbf{83.7}&\textbf{78.5}\\
         \hline
    \end{tabular}
    \caption{Results (AUROC $\uparrow$) of \textbf{conventional} OSR. The reported baseline results are from ~\cite{openood}.  OpenSlot achieves SOTA results on these conventional benchmarks~\cite{CIFAR10,tinyimagenet}.}
    \label{tab:conventional_osr_results}
\end{table}

\subsection{Ablation studies}
\label{sec:ans_abl}
This section tests the feasibility of OpenSlot. First, we display the semantic misalignment problem during slot predictions and discuss the effect of our ANS. Finally, we ablate the scoring functions, aggregation schemes, and hyperparameters.

\noindent\textbf{The semantic misalignment problem.} Using object-centric learning, we represent image semantics with a fixed set of slot features, each producing a class prediction. However, noise (background \& invalid) slots exist in every image and are likely to be incorrectly matched with the true class labels during Hungarian matching. We observe that the slots contributing to class predictions tend to represent noise rather than class objects. We call this misalignment between predictive slots and true class semantics “semantic misalignment".

\begin{figure}[t]
\begin{center}
\includegraphics[width=8.35cm]{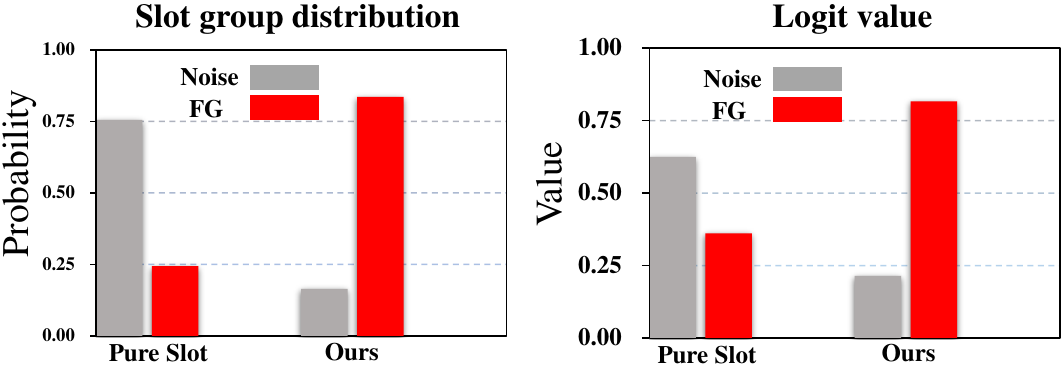}
\end{center}
   \caption{To compare OpenSlot with the Pure Slot baseline, we divide all slots into the foreground (FG) and Noise groups based on the semantics denoted by each slot. We then show the distribution of the max-logit slot (with the highest logit value from $\varphi_{fg}$) across these two groups (\textbf{Left}). Besides, we present each group's logit norm (\textbf{right}) (after Min-max normalization).  }
   \label{fig:slot_comparison}
\end{figure}

To show the threat of this problem, we consider the single-label classification task as an example; its class prediction is determined by the max-logit slot, the slot with the highest logit from the main classifier $\varphi_{fg}$. We utilize the ground-truth (GT) segmentation masks to label the object semantics captured by each slot. Specifically, we compute the valid semantic region for all slots (using attention masks) and select the one with maximum overlap with GT as the foreground (FG) slot, i.e., the true class object; while the remaining slots are labeled as Noise. In \refFig{fig:slot_comparison} (left), we display the distribution of the max-logit slot across the FG and Noise groups. The right side of \refFig{fig:slot_comparison} shows the average logit norms of these two groups.

\begin{figure}
    \centering
    \includegraphics[width=8.5cm]{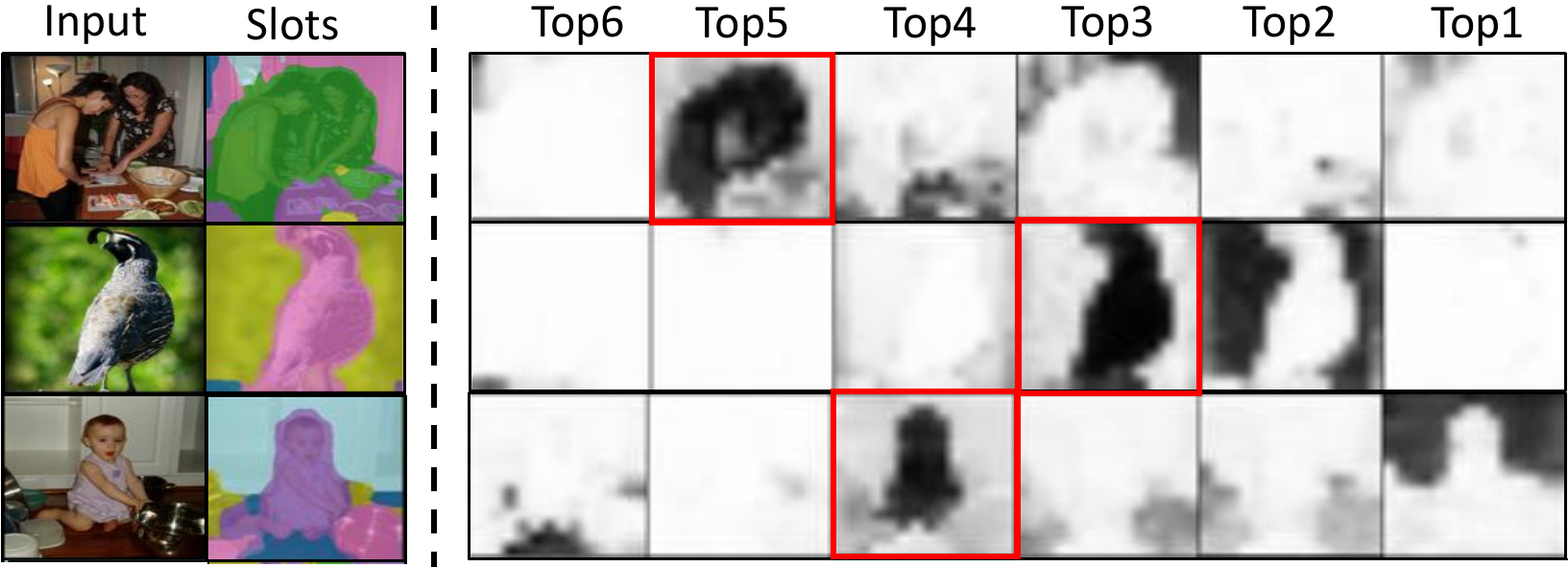}
    \caption{Display of the semantic misalignment problem. After pre-training, we learn slots for semantic representation (in different colors). In Pure Slot, we train the main classifier $\varphi_{fg}$ with vanilla Hungarian matching. When sorting (ascending) all slots by $\varphi_{fg}$'s output and displaying their attention masks (Top 6 $\rightarrow$ Top 1), we find that the noise (invalid \& background) slots have larger logit values than the true class slots (marked with red), indicating that $\varphi_{fg}$ is confused by noise slots.   }
    \label{fig:original_attention_masks}
\end{figure}

When employing the naive approach (Pure Slot, described in Section \ref{sec:method}), we observe that the max-logit slot is prone to falling into the Noise group. Besides, by sorting the slots in ascending order based on their logit values from $\varphi_{fg}$ and visualizing their attention masks in  \refFig{fig:original_attention_masks}, we note that the noise slots have larger logit values than the FG slots, consistent with the distribution shown in \refFig{fig:slot_comparison}.

These observations indicate the Pure Slot method (without ANS) can be easily confused by noise slot features in the classification training, failing to capture the true class semantics.

\noindent\textbf{Effects of ANS.} By comparison (\refFig{fig:slot_comparison}, ours), ANS ensures that the max-logit slot falls into the FG group, aligning with the true class semantics. Moreover, ANS increases the logit value gap between the FG and Noise slot groups.

\begin{figure*}
    \centering
    \includegraphics[width=16.5cm]{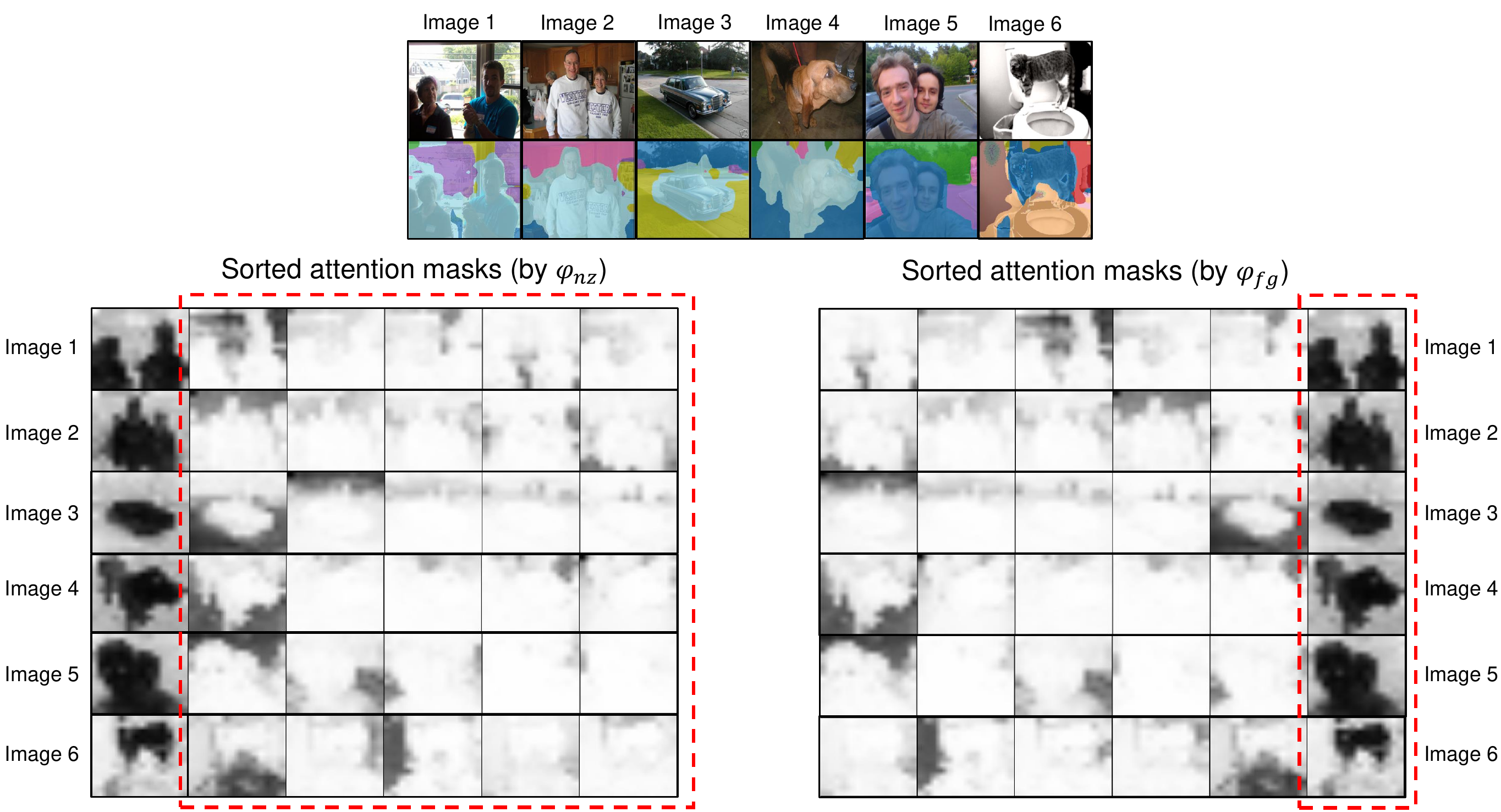}
    \caption{Demonstration of ANS' effects on slot separation. Our ANS excludes noise slots from Hungarian matching and enlarges their logit gap against the true class slots during training. In this example, we sort (ascending) attention masks based on slots' logit from the main classifier $\varphi_{nz}$ and the noise classifier $\varphi_{fg}$, respectively. In the right figure, the true class slots (cropped part) get the largest logit values. Meanwhile, $\varphi_{nz}$ (the left figure) can discriminate noise slots (cropped part).  }
    \label{fig:ANS_attention_masks}
\end{figure*}

Meanwhile, we perform a similar sorting and visualization analysis of the attention masks in \refFig{fig:ANS_attention_masks}. On the right side, the main classifier $\varphi_{fg}$ suppresses noise and assigns the slot corresponding to the true class object with the largest logit value. This finding aligns with the results shown in \refFig{fig:slot_comparison} (ours). On the left, the slots sorted by the outputs of the noise classifier $\varphi_{nz}$ clearly demonstrate its ability to effectively discriminate noise features.

\refFig{fig:slot_comparison} and the comparison between \refFig{fig:original_attention_masks} and \ref{fig:ANS_attention_masks} showcase that our ANS method effectively separates the noise slots and excludes them from semantic classification. During inference, when computing OSR scores for positive and negative images, the noise slots from both images produce similar values. This ensures that the primary score gap between positive and negative images arises from the semantic differences in class objects, facilitating a clearer class decision boundary.

Finally, we compare Pure Slot and OpenSlot in \refTab{tab:ablation_ans}. Our consistent performance gains across both the $\mathcal{M}$ and $\mathcal{H}$ test sets verify that ANS enhances the ability of slots to detect both super-label and full-label shifts.

These qualitative and quantitative results demonstrate that our ANS effectively discriminates noise slot features through a two-classifier design, isolating them during classification training. This ensures that true class slots are accurately matched with their labels, mitigating semantic misalignments. As a result, OpenSlot more effectively classifies both known and unknown objects in multi-label and mixed scenarios, improving the detection of super-label shifts.

\noindent\textbf{Choice of scoring functions.} To obtain the OSR results, we score the slot-based class predictions. We evaluate various out-of-distribution (OOD) detection methods, including four logit-based: MSP~\cite{msp}, ODIN~\cite{odin-score}, Energy~\cite{energy_ood}, MaxLogit~\cite{closed}, as well as the feature-based Mahalanobis distance~\cite{m_score}. In \refTab{tab:different_ood_scores}, we compare their performances. When combined with OpenSlot, MaxLogit, and Energy~\cite{energy_ood} consistently outperform other scoring methods, yielding superior OSR results.

\noindent\textbf{Effects of score aggregation methods.} This part investigates the impact of score aggregation schemes on OSR results. We combine the scores from all (All) slots to represent the test sample's overall “knownness” level. Besides, we design a selective (Sel.) scheme as an alternative. Specifically, we utilize the noise classifier $\varphi_{nz}$'s predictions to identify the foreground (FG) slots (with scores below a threshold $\gamma$) and the noise slots. During inference, we only aggregate the scores of the determined FG slots. Scoring with the energy method~\cite{energy_ood}, we report the size of the FG group (as a percentage of the entire slot set) when using different $\gamma$ values, and their OSR results.

In \refTab{tab:aggregation_schemes}, we observe a minor performance (AUROC) gap (when $\gamma=0.75$) between two schemes on the mixed $\mathcal{M}$ set. Notably, the determined FG group comprises only around 30\% of the slot set. This observation suggests that the basis for OpenSlot's OSR decision stems from the detected semantic differences among the class objects, i.e., unknown unknown classes (UUCs), in the FG group rather than from noise slots. This finding further verifies ANS's effect on noise separation.

\noindent\textbf{Effects of hyperparameters.} Here we ablate two hyperparamers: the attention threshold $\alpha$ and noise threshold $\beta$. 

\refTab{tab:hyperparameter_ablation} reports the OSR results when setting $\alpha$ and $\beta$ of different values. We observe that OpenSlot reaches the best performances when $\alpha=0.50$ and $\beta=0.75$ on both benchmarks. Besides, these ablation results suggest that $\beta$ has a larger effect on OSR results than $\alpha$, showing the importance of noise discrimination in slot predictions.

\begin{table}
    \centering
    \small
    \begin{tabular}{c|c c | c c}
    \hline
     \multirow{2}{*}{Method}& \multicolumn{2}{c|}{VOC-6/14}& \multicolumn{2}{c}{COCO-20/60}\\
      ~&$\mathcal{H}$& $\mathcal{M}$&$\mathcal{H}$& $\mathcal{M}$ \\
        
    \hline
    Pure Slot &80.4 / 76.7&55.7 / 92.1&76.0 / 78.8&67.8 / 84.7\\
   \rowcolor{Gray} Ours &92.0 / 56.3&67.3 / 88.4&86.7 / 62.7&73.5 / 79.7\\

    \hline
    \end{tabular}
    \caption{Anti-noise-slot (ANS) evaluation. All results are AUROC $\uparrow$ / FPR@95 $\downarrow$.}
    \label{tab:ablation_ans}
\end{table}


\begin{table}
    \centering
    \scriptsize
    \begin{tabular}{c|c c | c c}
    \hline
     \multirow{2}{*}{OOD  Metrics}& \multicolumn{2}{c|}{VOC-6/14}& \multicolumn{2}{c}{COCO-20/60}\\
      ~&$\mathcal{H}$& $\mathcal{M}$&$\mathcal{H}$& $\mathcal{M}$ \\
        
    \hline
    Mahalanobis~\cite{m_score} &82.7 / 81.7&50.1 / 95.2&87.0 / 79.4&50.7 / 97.7\\
    MaxLogit~\cite{closed} &92.0 / 56.3&67.3 / 89.2&88.4 / \textbf{62.7}&73.5 / \textbf{79.7}\\
    MSP~\cite{msp}&89.1 / 69.1& 64.6 / 90.2&80.7 / 79.9&69.4 / 85.3\\
    Odin~\cite{odin-score} &92.0 / \textbf{51.5}& 66.4 / 88.5&84.2 / 73.0&69.7 / 83.2\\
    Energy~\cite{odin-score} &\textbf{93.2} / 53.3& \textbf{71.3} / \textbf{86.7}&\textbf{89.6} / 70.1&\textbf{74.9} / 83.1\\
    \hline
    \end{tabular}
    \caption{Ablation results (AUROC $\uparrow$ / FPR@95 $\downarrow$) of OpenSlot using different OOD detection functions. The Energy~\cite{energy_ood} and Maxlogit~\cite{closed} are better metrics for OpenSlot. }
    \label{tab:different_ood_scores}
\end{table}

\begin{table}
\small
    \centering
    \begin{tabular}{ccc|cc}
         \hline
         
         Scheme&$\gamma$&FG (\%)&$\mathcal{H}$&$\mathcal{M}$\\
         \hline
        \multirow{4}{*}{Sel.} ~&0.05&9.2&68.7 / 86.7&57.4 / 90.4 \\
          ~&0.25&25.9&84.2 / 74.1&64.6 / 88.1\\
 ~&0.50&27.0&86.3 / 73.8&65.6 / 87.6\\
  ~&0.75& 32.1&89.1 / 77.2&67.0 / 89.6\\
  \hline
 All&-&-&93.2 / 53.3&71.3 / 86.7\\
  
         \hline
    \end{tabular}
    \caption{Effects of $\gamma$. We test on the single-label VOC and use ~\cite{energy_ood} for OSR evaluation (AUROC $\uparrow$ / FPR@95 $\downarrow$)}
    \label{tab:aggregation_schemes}
\end{table}

\begin{table}
    \centering
    \small
    \begin{tabu}{>{\centering}m{0.45cm} >{\centering}m{0.45cm} |m {1.4cm}<\centering m {1.4cm}<\centering m {1.4cm}<\centering m {1.4cm}<\centering}
    \hline
    \multirow{2}{*}{$\alpha$} & \multirow{2}{*}{$\beta$} & \multicolumn{2}{c}{VOC-6/14}& \multicolumn{2}{c}{COCO-20/60}\\
      ~&~&$\mathcal{H}$& $\mathcal{M}$&$\mathcal{H}$& $\mathcal{M}$\\
    \hline
         \multirow{3}{*}{0.25}  &0.25 &71.1 / 80.9 &53.4 / 96.4 &69.1 / 82.7 &51.1 / 97.6   \\
         ~ &0.50 &84.3 / 70.5 &59.7 / 93.6 &77.8 / 75.4 &61.7 / 90.1  \\
         ~ &0.75 &89.1 / 62.5 &63.7 / 90.5 &82.9 / 69.4 &70.7 / 85.8   \\
         \hline
         \multirow{3}{*}{0.50}&0.25 &76.6 / 79.7 &56.1 / 94.7 &77.1 / 80.6 &63.1 / 91.8 \\
         ~ &0.50 &82.7 / 75.4 &59.1 / 92.3 &78.4 / 74.5 &64.7 / 87.8 \\
    \rowcolor{Gray}
    ~ &0.75&\textbf{92.0 / 56.3}&\textbf{67.3 / 88.4}&\textbf{86.7 / 62.7}&\textbf{73.5 / 79.7} \\
    \hline
        \multirow{3}{*}{0.75}&0.25 &76.8 / 77.4 &62.5 / 91.7  &76.1 / 76.4 &65.9 / 91.7  \\
         ~ &0.50 &84.9 / 70.8 &62.9 / 90.4 &80.2 / 70.3 &69.8 / 87.5  \\
         ~ &0.75 &90.9 / 60.1 &65.8 / 89.3 &85.1 / 64.2 &72.7 / 80.9  \\
         \hline
    \end{tabu}
    \caption{Effects (AUROC $\uparrow$ / FPR@95 $\downarrow$) of the attention threshold $\alpha$ and the noise threshold $\beta$. We get the best results when setting $\alpha$ and $\beta$ to 0.50 and 0.75, respectively.}
    \label{tab:hyperparameter_ablation}
\end{table}

\subsection{Effect of slot prediction in OSR scoring} 
\label{sec:scoring_comparison}
In this section, we compare OpenSlot with conventional OSR models to test the effectiveness of slot-based predictions in addressing the mixed OSR problem. We do this by analyzing the differences in their respective scoring mechanisms.

Given an input image $I\in \mathbb{R}^{ H \times W \times 3}$ with multiple class semantics, a conventional OSR approach (either with the CNN/ViT) is to use classifier $f$ maps the semantic features of $I$ to $\hat{l}\in \mathbb{R}^{|\mathcal{K}|}$ and get $|\mathcal{K}|$ (the number of KKCs) logit values. Based on $\hat{l}$, we compute a deterministic score for OSR evaluation.  By contrast, our method uses the classifier $f_{slot}$ across slots and gets $\hat{l}_{slot}\in\mathbb{R}^{N\times|\mathcal{K}|}$, where each slot has an independent class prediction for the denoted object semantic.  

\begin{figure}
    \centering
    \includegraphics[width=8.2cm]{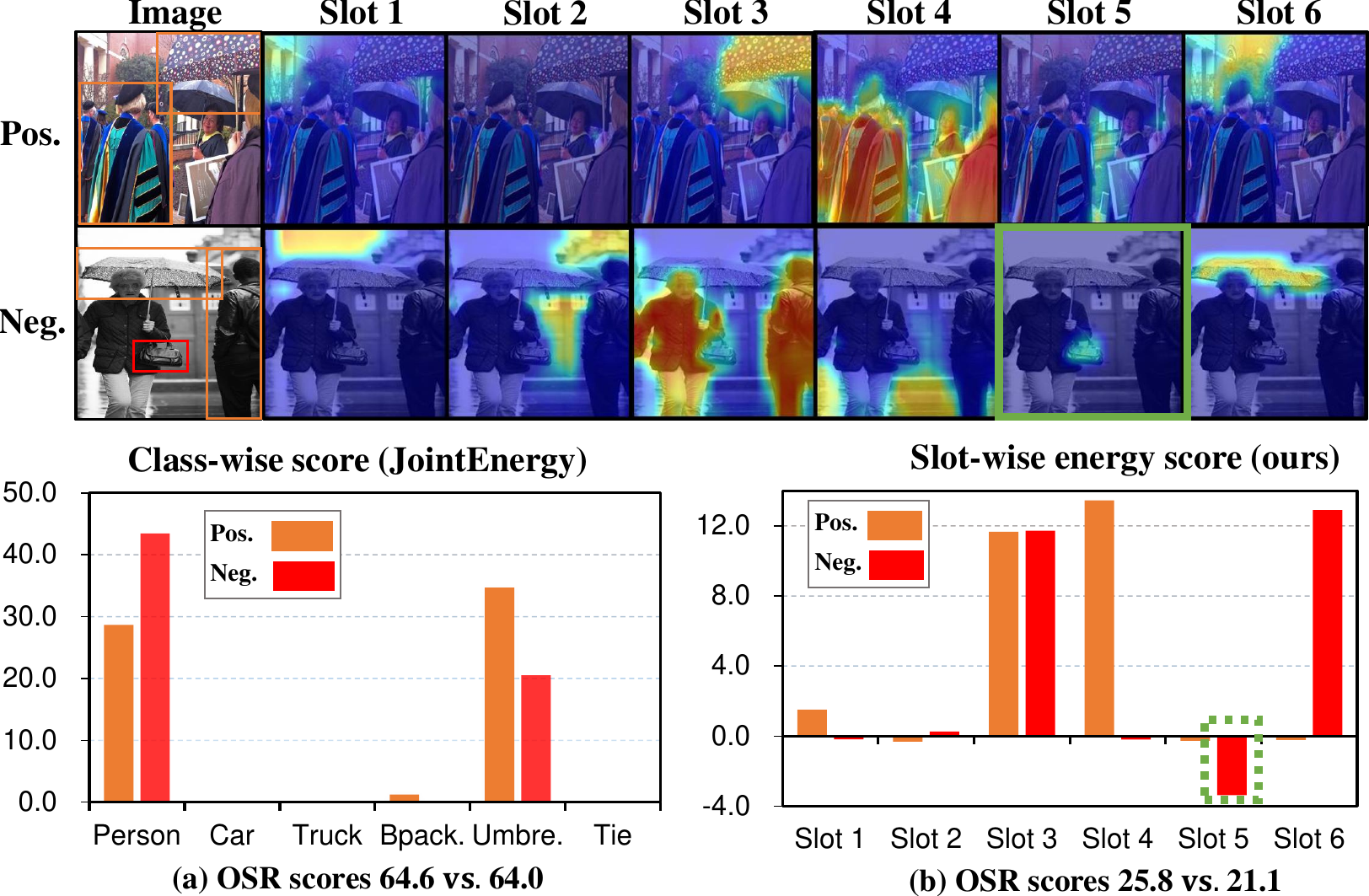}
    \caption{To compare with the conventional OSR approach, we input the positive (Pos.) and the mixed negative (Neg.) samples into ResNet50~\cite{multi_label_ood} and OpenSlot. (a) we compute JointEnergy~\cite{multi_label_ood} of two samples and show class-wise values. The positive and the negative image score 64.6 and 64.0, respectively. The known class objects (“Umbre.” and “Person”) in the negative image have high scores in corresponding class channels, making their overall score value almost identical to the positive image. By contrast (b), we aggregate the slot prediction score ~\cite{energy_ood} and obtain 25.8 vs. 21.1 for two samples. Combined with attention masks, we show that the unknown slot (“handbag”, slot 5) gets the lowest score, which provides meaningful discrimination information for OSR decisions.  }
    \label{fig:cnn_comparison}
\end{figure}

\refFig{fig:cnn_comparison} presents an example comparison between JointEnergy~\cite{multi_label_ood} and our OpenSlot, when feeding a positive and a mixed negative image. JointEnergy~\cite{multi_label_ood} displays nearly no difference in the scores between the two samples (64.6 $ vs. $ 64.0). It is because its $\hat{y}$ is built upon the global semantic content of the image $I$, which is largely affected by the prominent known classes. The label shift of the inconspicuous unknown class (“handbag”) is covered by high scores in known class channels (\refFig{fig:cnn_comparison}, bottom left), making the negative indistinguishable from the positive in terms of score values. 

By contrast, our slot approach addresses the threat of known classes in the mixed OSR problem by producing independent class predictions for captured image semantics. During inference,  every slot prediction in $f_{slot}$ is measured with the score function and is counted for OSR evaluation, without effects from others. In \refFig{fig:cnn_comparison} (bottom right), we observe that the score gap between the positive and the negative image primarily comes from the detected unknown class object.

\noindent\textbf{Limitations.} Although exceeding existing OSR studies in detecting the super-label shift, OpenSlot underperforms the best method when handling the full-label shift in the multi-label setting. We attribute this to our model's inferior classification accuracy in the closed set, and consider that other matching/cost computation methods~\cite{set_prediction} could be potential solutions, which we leave for future investigation.

\begin{figure}
    \centering
    \includegraphics[width=8cm]{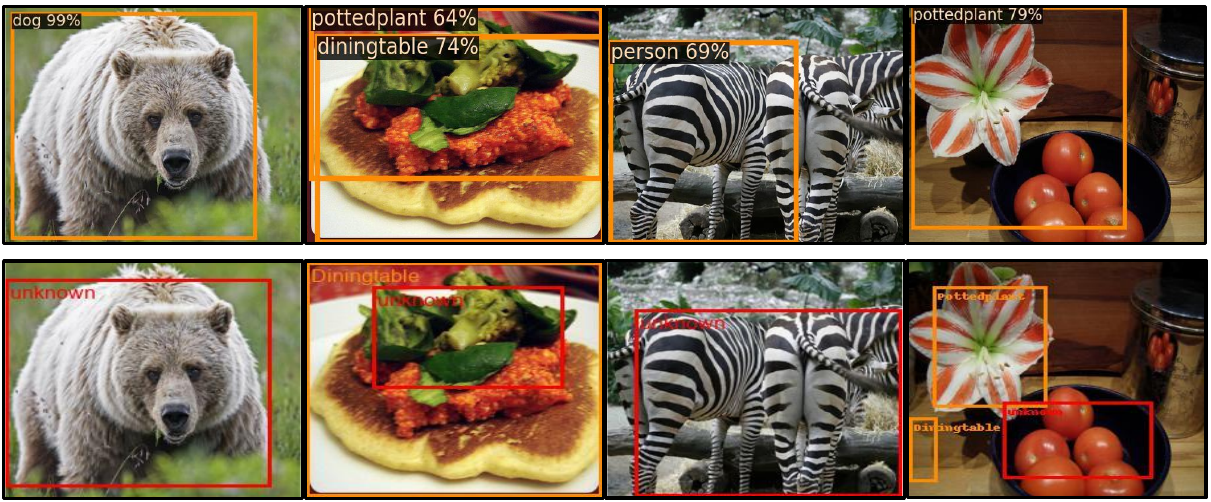}
    \caption{Object detection results on open-set images by Faster-RCNN~\cite{faster_rcnn} (the top row) and ours (the bottom row). The close-set data is VOC. \textbf{Orange}: detected objects and classified as known \textbf{Red}: detected unknown objects. In this example, we correct the misprediction of Faster-RCNN and successfully localize unknown class objects.    
    }
    \label{fig:osod_correction}
\end{figure}
\begin{figure}
    \centering
    \includegraphics[width=8cm]{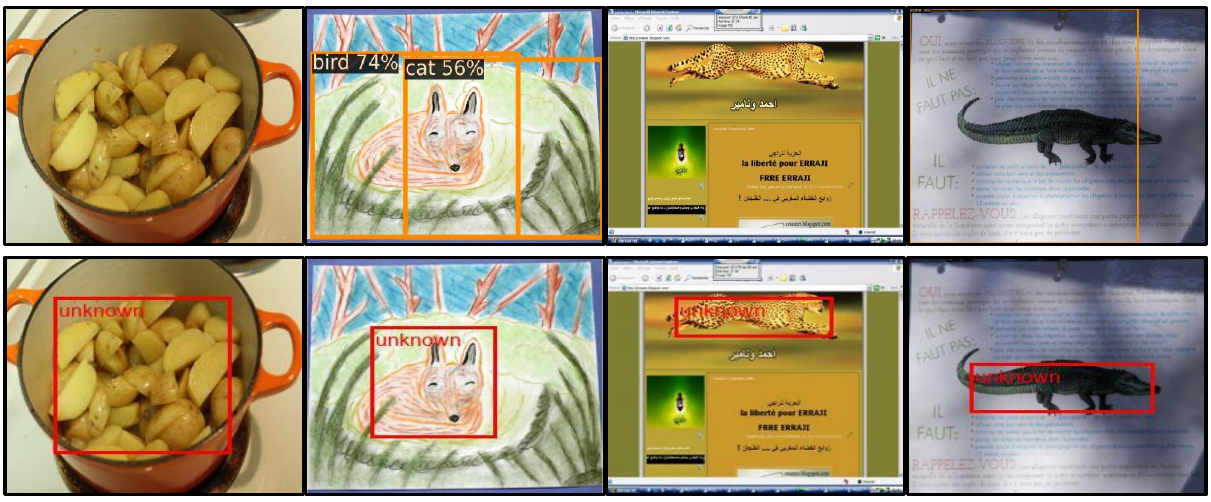}
    \caption{Object detection results on open-set images with domain gap from the close-set data: Faster-RCNN (top) and ours (bottom). Faster-RCNN is constrained by the predefined classes in the close-set and fails to generate meaningful object proposals. By contrast, OpenSlot generates reliable objects of interest and distinguishes the known and unknown.
    }
    \label{fig:osod_discovery}
\end{figure}
\begin{table}[t]
    \centering
    \small
    \begin{tabular}{c<{\centering}  c<{\centering}c<{\centering}}
    \hline
    Method & COCO & OpenImages \\
    \hline

    MSP~\cite{msp} & 83.5 / 71.0 & 81.9 / 73.1 \\
    ODIN~\cite{odin-score} & 82.2 / 59.8 & 82.6 / 63.1 \\
    KNN~\cite{knn} & 87.1 / 52.7 & 84.5 / 53.7 \\
    Energy score~\cite{energy_ood} & 83.7 / 56.9 & 83.0 / 58.7 \\
    VOS~\cite{vos} & 88.7 / 47.5  & 85.2 / 51.3\\
    SIREN-vMF~\cite{siren} & 85.4 / 64.7 &  82.8 / 68.5 \\
    \rowcolor{Gray}Ours & 85.2 /  66.3 &83.4 / 61.8\\
    
    \hline
    \end{tabular}
    \caption{Open-set object detection results (AUROC $\uparrow$ / FPR@95 $\downarrow$). Without using bounding boxes in training, our method outperforms several detector-based studies.}
    \label{tab:osod_detetection}
\end{table}

\subsection{Yielding Benefits to Open-Set Object Detection} 
\label{sec:osod}
Apart from providing a binary decision on known and unknown, OpenSlot can also localize unknown class objects within images to explicitly explain the label shift, thus achieving the effect of open-set object detection (OSOD).

 Following the setup in~\cite{vos}, we choose VOC as the close-set data and test on open-set images from COCO and OpenImages~\cite{openimages}. During training, OpenSlot only uses class labels.

\noindent\textbf{Results.} Table \ref{tab:osod_detetection} compares OpenSlot with existing OSOD studies. Without employing a Region Proposal Network~\cite{faster_rcnn}, OpenSlot even outperforms several object detector-based approaches and achieves performance comparable to SIREN-vMF~\cite{siren} (Detailed localization results are reported in Appendix).  \refFig{fig:osod_correction} presents example results where OpenSlot corrects the mispredictions of the detector.

Notably, unlike traditional methods that rely on predefined classes and selective search algorithms for proposal generation, OpenSlot discovers objects from image data in an unsupervised manner~\cite{Dino,dinosaur}, and can obtain robust object representations against domain shifts~\cite{generalization_ocl}. When tested on open-set images with significant domain differences from the closet-set (\refFig{fig:osod_discovery}), these properties ensure that OpenSlot identifies different objects of interest, leading to improved localization of unknown objects and reflecting our slot approach's stronger generalization ability than detector-based methods.

\noindent\textbf{Overhead comparison.} \refTab{tab:comparing_inference_cost} compares the computational cost of OpenSlot, Faster R-CNN (ResNet50)\cite{faster_rcnn}, and DETR (ResNet50)\cite{DETR} in terms of model size, training time, and inference speed. Our OpenSlot features a lightweight design with only 2.4 million parameters (including the pre-training and classification stages), requiring just 3.2 hours for training. This is significantly lower than existing CNN/transformer-based object detection models. Additionally, OpenSlot demonstrates faster inference times for object detection, underscoring its efficiency advantages.
\begin{table}
\small
    \centering
    \begin{tabular}{c|ccc}
         \hline
         Method&Param. &Train. time&Infer. speed\\
         \hline
         Faster-RCNN~\cite{faster_rcnn}&33.0 M& 3.7 h & 0.24 s\\
         DETR~\cite{DETR}&41.6 M& 3.9 h & 0.29 s\\
         \rowcolor{Gray}Ours& 2.4 M& 3.2 h&0.16 s \\ 
         \hline
    \end{tabular}
    \caption{We compare Faster-RCNN and OpenSlot on RTX 4090, in terms of the parameter size (million, M), training time (hours, h), and inference speed (second/per image).}
    \label{tab:comparing_inference_cost}
\end{table}

%% file: sec/Conclusion.tex
This paper introduces the mixed open-set recognition (OSR) problem, which considers the joint occurrence of known and unknown class objects in negative images. To address this challenge, we present the OpenSlot framework, built upon the object-centric learning approach. Through the proposed anti-noise-slot (ANS) technique, we effectively distinguish noisy slots from true class features and exclude them from the classification process, thus addressing the semantic misalignment issue inherent in slot-based class predictions. Extensive experiments and ablation studies demonstrated that OpenSlot significantly outperforms existing studies in handling the mixed OSR problem and achieves SOTA results in conventional OSR benchmarks. Additionally, OpenSlot can be extended to open-set object detection, localizing unknown class objects and explicitly explaining label shifts. The competitive results and computational efficiency show the advantages of our model.

%% file: sec/Appendix.tex
In this appendix, we introduce the detailed benchmark information and implementations. 

\subsection{Mixed OSR}
Pascal Voc (VOC) supports classification (VOC-Cls), segmentation (VOC-Seg). For the evaluation convenience, we use VOC-Seg in  \refTab{tab:single_label_osr} and use the original COCO dataset. 

\noindent{\textbf{setup.}} We merge each dataset's training and validation set and perform the class split. First, we make a statistic about the object class distribution and find 6 VOC\footnote{“person”, “bird”, “car”, “cat”, “dog”, “person”} and 20 COCO classes\footnote{“airplane”, “train”, “boat”, “traffic light”, “fire hydrant”, “stop sign”, “bench”, “bird”, “cat”, “horse”, “sheep”, “cow”, “elephant”, “bear”, “zebra”, “giraffe”, “toilet”, “clock”, “vase”, “ted. bear”} have over 450 single-label images, which we consider as KKC candidates; the rest of the classes are used as UUCs. 

For single-label tasks, we use all KKC and UUC candidates to construct VOC-6/14 and COCO-20/60. We use VOC-Cls and COCO to conduct the multi-label experiments and build up the test sets with the same criteria as the single-label setting. In COCO-{20/60, 40/40, 60/20} tasks, KKCs are randomly sampled, and the results vary with the sampling effects.

\subsection{Experiment details}
\noindent\textbf{Mixed OSR.} In single-label experiments, we set the learning rate (LR) as 0.0004 (half it every 40 epochs) and train the OpenSlot model for 200 epochs. We set lr to 0.005 and train 250 epochs in the multi-label case.  Besides, we add a loss weight $\lambda$ before the noise loss $L_{nz}$ (Eq. \ref{eq:overall_loss}) and set it to 0.01. 

\noindent\textbf{Open-set object detection.}  We use the noise classifier $\varphi_{nz}$'s predictions and a threshold (0.75) to determine the foreground (FG) slots (below threshold). We then mask (over 0.5) the pixels in the FG attention mask and consider the resulting binary mask as the slot's semantic region. Assuming a meaningful semantic object must be over 40 pixels, we obtain a valid object mask to compute bounding boxes. Finally, we score~\cite{energy_ood} all FG slots and consider those with scores over five as known objects and the rest as unknown.

\begin{table*}[t]
\centering
\scriptsize

\begin{tabu}{m{0.5cm}<{\centering}|m{0.37cm}<{\centering}m{0.37cm}<{\centering}m{0.37cm}<{\centering}m{0.37cm}<{\centering}m{0.37cm}<{\centering}m{0.37cm}<{\centering}m{0.37cm}<{\centering}m{0.37cm}<{\centering}m{0.37cm}<{\centering}m{0.37cm}<{\centering}m{0.37cm}<{\centering}m{0.37cm}<{\centering}m{0.37cm}<{\centering}m{0.37cm}<{\centering}m{0.37cm}<{\centering}m{0.37cm}<{\centering}m{0.37cm}<{\centering}m{0.37cm}<{\centering}m{0.37cm}<{\centering}m{0.37cm}<{\centering}}
    \tabucline[1pt]{-} 
\textbf{mAP}&\textbf{aero}&\textbf{bicy}&\textbf{bird}&\textbf{boat}& \textbf{bottle}&\textbf{bus}&\textbf{car}&\textbf{cat}&\textbf{chair}&\textbf{cow}&\textbf{dt.}&\textbf{dog}&\textbf{horse}&\textbf{motor}& \textbf{pers.}&\textbf{pott}&\textbf{sheep}&\textbf{sofa}&\textbf{train}&\textbf{tv.}\\
\tabucline[1pt]{-} 

30.8&43.9&56.4&35.5&26.0&14.0&66.5&53.8&32.6&7.1&36.7&16.2&36.7&38.7&52.5&8.2&12.7&36.2&42.9&55.9&50.9\\
\tabucline[1pt]{-} 
\end{tabu}
\caption{Category-level $AP_{50}$ results on Pascal Voc 2007. }
\label{tb:perclass_pascal_segmentation}
\end{table*}